# InFER: A Multi-Ethnic Indian Facial Expression Recognition Dataset

Syed Sameen Ahmad Rizvi[a], Preyansh Agrawal[b], Jagat Sesh Challa[c] and Pratik Narang[d]

*Department of CSIS, BITS Pilani, Pilani Campus, Rajasthan, India*
{*p20190412, f20190052, jagatsesh, pratik.narang*}@*pilani.bits-pilani.ac.in*

Abstract: The rapid advancement in deep learning over the past decade has transformed Facial Expression Recognition (FER) systems, as newer methods have been proposed that outperform the existing traditional handcrafted techniques. However, such a supervised learning approach requires a sufficiently large training dataset covering all the possible scenarios. And since most people exhibit facial expressions based upon their age group, gender, and ethnicity, a diverse facial expression dataset is needed. This becomes even more crucial while developing a FER system for the Indian subcontinent, which comprises of a diverse multi-ethnic population. In this work, we present InFER, a real-world multi-ethnic Indian Facial Expression Recognition dataset consisting of 10,200 images and 4,200 short videos of seven basic facial expressions. The dataset has posed expressions of 600 human subjects, and spontaneous/acted expressions of 6000 images crowd-sourced from the internet. To the best of our knowledge InFER is the first of its kind consisting of images from 600 subjects from very diverse ethnicity of the Indian Subcontinent. We also present the experimental results of baseline & deep FER methods on our dataset to substantiate its usability in real-world practical applications.

## 1 INTRODUCTION

Facial expression is one of the most prominent and powerful means by which human beings convey their emotional states and intentions(Darwin and Prodger, 1998; Tian et al., 2001). Studies have shown that 55% of the messages pertaining to feelings are in the facial expressions, 7% comprises spoken words, and the rest 38% belongs to para linguistics (i.e., the style in which verbal communication happens)(Mehrabian and Russell, 1974). Hence, facial expressions have proven to play a pivotal role in the entire information exchange process. (Ekman, 1994), on the basis of a cross-cultural study(Ekman and Friesen, 1971), defined six basic sets of emotions to be the standard way humans perceive certain basic emotions regardless of their culture. These pivotal emotions include anger, fear, disgust, happiness, sadness, and surprise. Over the last two decades, with the dramatic development in artificially intelligent systems, numerous attempts have been made to automatically detect facial expressions due to its versatile scope in sociable robots, medical diagnosis, surveillance systems, behavioral research, and many other human-computer interaction systems. Facial Expression Recognition (FER) systems can broadly be categorized into two types: traditional and deep FER (Li and Deng, 2020). In the early stages, several traditional handcrafted methods like local binary patterns and non-negative matrix factorization were used. As we evolved in the field of deep neural networks, various methods were proposed that used paradigms such as convolution neural networks, deep belief networks, generative adversarial networks, and recurrent neural networks.

Deep learning refers to the process of extracting high-level abstractions through multiple layers of non-linear transformation and representation. Such a supervised learning approach would always require a sufficiently large training dataset covering all the scenarios. Since most people, based upon age group, gender, ethnicity, and other socio-cultural factors, exhibit facial expressions differently, for a deep FER system to scale up to such a diverse audience requires

---

[a]  https://orcid.org/0000-0002-3919-5074
[b]  https://orcid.org/0000-0002-5232-6446
[c]  https://orcid.org/0000-0002-9794-0087
[d]  https://orcid.org/0000-0003-1865-3512

This research is supported by grants from Kwikpic.

Table 1: An overview of existing facial expression datasets. [P = posed; S = spontaneous]

| Dataset | Samples | Subject | Condition | Elicitation | Expression Distribution |
|---|---|---|---|---|---|
| CK+ | 593 image sequences | 123 | Lab | P & S | seven basic expressions contempt |
| MMI | 740 images 2900 videos | 25 | Lab | P | seven basic expressions |
| JAFFE | 213 images | 10 | Lab | P | seven basic expressions |
| TFD | 112,234 images | N.A. | Lab | P | seven basic expressions |
| FER-2013 | 35,887 images | N.A. | Web | P & S | seven basic expressions |
| AFEW 7.0 | 1,809 images | N.A. | Movie | P & S | seven basic expressions |
| SFEW 2.0 | 1,766 images | N.A. | Movie | P & S | seven basic expressions |
| Multi-PIE | 755,370 images | 337 | Lab | P | smile, surprised, squint, disgust, scream, and neutral |
| BP4D | 328 2-D & 3-D videos | 41 | Lab | P | seven basic expressions |
| Oulu-CASIA | 2,880 image sequences | 80 | Lab | P | six basic expressions without neutral |
| ExpW | 91,793 images | N.A. | Web | P & S | seven basic expressions |
| ISED | 428 videos | 50 | Lab | S | Happiness, surprise, sadness, disgust. |
| EmotioNet | 1,000,000 images | N.A. | Web | P & S | 23 basic and compound emotions |
| RAF-DB | 29,762 images | N.A. | Web | P & S | seven basic expressions 12 compound expressions |
| AffectNet | 450,000 images | N.A. | Web | P & S | seven basic expressions |

a diverse dataset. Moreover, since most datasets currently used are generated in a lab-based controlled environment, they usually contain high-quality images with frontal head poses. In the real-world scenarios, there are many more challenges such as non-frontal head poses, occlusions, and poor illumination. Deploying a deep neural network trained on such a dataset yields lower levels of performance and accuracy.

The Indian subcontinent, known for its rich socio-cultural composition, encompasses a variety of ethnicities. A study done on the ethnic composition of Indian people enlists a variety of ethnic groups, including Dravidian, Indo-Aryan, Mongoloid, Aryo-Dravidian, Monglo-Dravidian, Scytho-Dravidian and Turko-Iranian (Risley, 1891). Therefore deploying a real-world practical FER application on such a diverse population without a dataset adequately representing all ethnic and age groups is a challenging task. Henceforth, for such a heterogeneous population, there is a need for a diverse dataset that can cater to the diversified requirements of the Indian subcontinent, a home to more than 1.4 billion people.

In this paper, we propose a diverse India-specific real-world - in the wild facial expression dataset - InFER, which aims to cover various ethnicities and age groups of the Indian subcontinent. The major contributions of this work include:

- We created InFER dataset that consists of seven basic expressions of anger, disgust, fear, happiness, neutral, sadness, and surprise, of 600 human subjects of varied Indian ethnicity, age groups, and gender along with their respective labels.

- It comprises of 10,200 images and 4,200 short videos, of which 4,200 images and short videos belong to 600 human subjects, and the remaining 6000 images correspond to real-world facial expression images taken from online sources.

- We show the performance analysis on baseline traditional as well as recent state-of-the-art deep FER models.

To the best of our knowledge, our dataset is the first of its kind to contain the largest collection of human subjects in the facial expression recognition domain. This is also the first multi-ethnic diverse India-specific dataset presented in the domain of FER.

## 2 Related Work

A comprehensive dataset with adequate labeled training data covering as many variations in terms of ethnicities, age group, gender and environments is crucial for designing a deep FER system. Table 1 provides an overview of the most common existing facial expression datasets. (Li and Deng, 2020). People with diverse ethnicity, age group, and gender exhibit facial expressions differently; the absence of ethnically rich & diverse dataset is the major limitation among the existing datasets. This limitation deepens in a country like India, which has a diverse population. Moreover, dataset bias and imbalance distribution are other common issues with FER datasets that results from the practicality of sample acquirement. Expressions like happiness & anger are comparatively easier to capture compared to expressions like disgust & fear, which are less common.

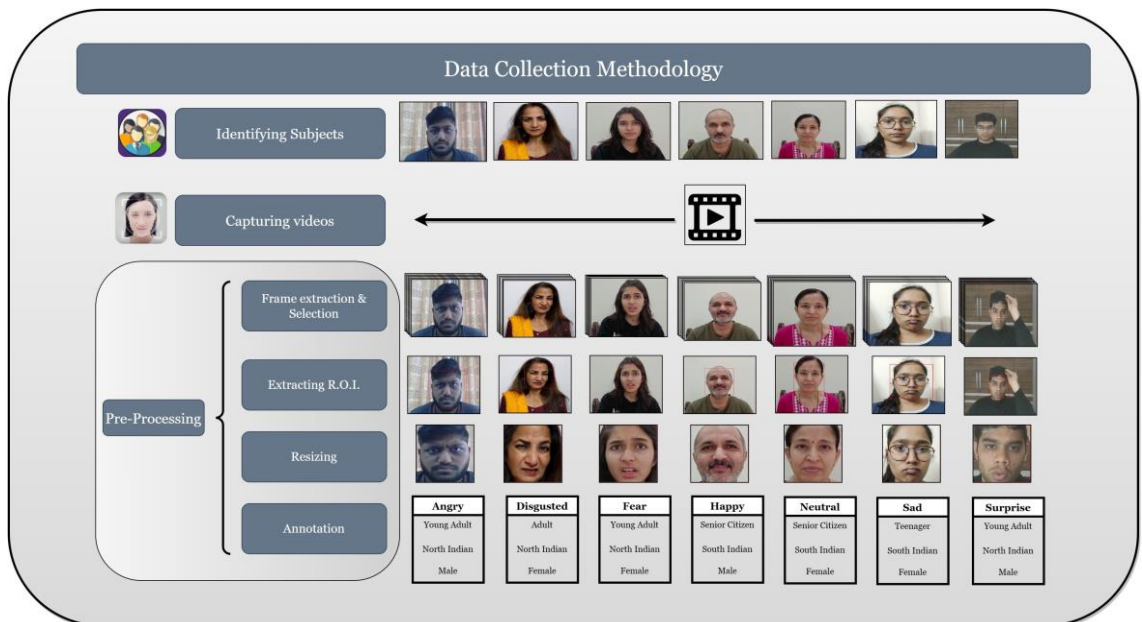

Figure 1: An overview of data collection methodology for human subjects.

## 3 Data Collection Methodology

This section describes the methodology employed to create the InFER dataset. The proposed dataset comprises of two segments: (*i*) posed emotions from 600 human subjects; (*ii*), spontaneous /acted emotions extracted from online sources, which includes 6000 collected on a crowd-sourced basis by an image search API. The methodology for both segments are discussed separately.

### 3.1 Human Subjects

Posed emotions can only be captured by consent and voluntary participation of the subjects. For this purpose, we constituted a team of 10 members who identified and persuaded human subjects to participate in our data collection task. All these members were given a thorough understanding of data collection strategy, expression styles, illumination conditions, variance in terms of gender, age group, and ethnicity, and lastly, the outlier scenarios such as non-frontal head posed and occlusions, i.e., presence of spectacles, beard, mustache that can disguise some vital information pertaining to facial expressions. Since our focus here is to develop a real-world dataset, there was no specific lab setup for capturing posed human expressions. The motivation behind the fore-said approach is that we wanted our dataset to scale for real-world applications. Often, lab-based environments lack real-world obstructions, occlusions, and outlier scenarios. The methodology adopted to capture posed human facial expression is a 3 stage process illustrated in Fig.1. We shall discuss each stage in detail.

**Subject Identification**: Each member of our data collection team was assigned to gather expressions of 60 human subjects. Emphasis was on maximizing the variation in terms of ethnicity, age group, and gender. For this purpose, we categorized the country into five zones: north, east, west, south, and northeast. Adequate representation from each zone would help us gather an ethnically rich dataset. It was decided that representation of each zone must be at least 15% and at most 25% for each member of our data collection team. Similarly we divided age groups into five classes, i.e. {5-10 }children, {11-19 }teenagers, {20-29} young-adults, {30-59 }adults, {60+ }senior citizens. A similar kind of proportion was maintained for age groups so that each age class has adequate representation. Regarding gender, an almost equal proportion of subjects was expected amongst males and females. Due consent from all the participants or their guardians (in the case of minors) was signed for agreeing to use their facial images for research purposes.

**Capturing Videos** : In this stage, a 5-10 second video of the concerned subject was captured for all seven basic expressions. The facial expression started from a neutral expression, went to the peak of the expres-sion, and then came back to a neutral expression. Factors like proper illumination conditions, the distance between the camera and subject (ideally kept between 2-3 feet), and a few outlier scenarios like beard mustache, turban, and spectacles were also incorporated in the process.

**Pre-processing**: In pre-processing firstly the video file is parsed into its respective frames, and the frame that best represents the peak of the facial expression is selected. Secondly the region of interest (ROI) i.e. human face, is extracted using Haarcascade face detector (Viola and Jones, 2001). Thereafter, all the extracted facial images representing the respective peak of the emotion are resized to 256X256 pixels. Finally, each facial image is labeled with its respective expression type, age class, ethnic class, and gender type.

### 3.2 Online Sources

Posed expressions may not always scale up to real-world applications. Therefore, a substantial percentage of the dataset must contain acted or spontaneous facial expressions. For this purpose, we employ a crowd-sourcing strategy to gather facial expressions from various online sources. Search keywords and an image search API facilitate us in collecting diverse images with facial expressions from varied online sources. Since the API we used yielded results in a well-structured XML format, where the URLs can be conveniently parsed, a set of keywords was used to search for images with facial expressions.Subsequently, a set of keywords were used for specifically curated search queries to collect facial expression images with a balanced distribution in terms of expressions, ethnicity, age group, and gender. These keywords with respect to subjects included (smile, joyful, giggle, cry, upset, anger, rage, fearful, scared, frightened, terrified, shocked, astonished, disgust, and expressionless). The keywords employed for maintaining ethnic, age, and gender balance included - child, adolescent, adult, senior-citizen, boy, man, male, girl, lady, female, punjabi[1], marathi, tamil, telugu, bengali, north-Indian, south Indian, north-eastern Indian. A total of 6000 such facial expression images were collected and are presented in our dataset.

**Dataset Annotation**: Annotating 6000 images is a time-consuming and challenging task. Considering this, a team of 5 annotators was assigned to annotate all the 6000 images, with their respective age, gender, ethnicity, and expression labels. Each annotator was instructed and trained with the psychological knowledge of seven basic facial expressions through a demo exercise of annotating 1500 facial expression images from some of the already existing datasets (Table 1). Since these five different annotations would have the respective biases associated with individual annotators, we calculate a confidence matrix, where each annotator had to review and rate the annotation of the other four annotators between 0 to 1. The annotated data was provided to them anonymously so as to maintain confidentiality and have a fair review and assessment of annotations. The confidence matrix was calculated for each emotion. Table 2 shows a confidence matrix for happiness expression, where each row represents cross-confidence scores given by the other four annotators.

---
[1]Punjabi,Marathi, Tamil, Telugu, Bengali are the people belonging to the of Punjab, Maharshtra, Tamil Nadu, Andhra Pradesh, Bengal respectively; federal states of Union of India.

Table 2: Confidence matrix for Happiness expression.

| | | Scores | | | |
|---|---|---|---|---|---|
| | Annotator 1 | Annotator 2 | Annotator 3 | Annotator 4 | Annotator 5 |
| Annotator 1 | $x$ | 0.85 | 0.82 | 0.87 | 0.83 |
| Annotator 2 | 0.84 | $x$ | 0.79 | 0.81 | 0.80 |
| Annotator 3 | 0.91 | 0.89 | $x$ | 0.92 | 0.88 |
| Annotator 4 | 0.85 | 0.78 | 0.75 | $x$ | 0.80 |
| Annotator 5 | 0.90 | 0.86 | 0.85 | 0.84 | $x$ |

Cumulative confidence scores (C.C.Score) are calculated for each row using equation 1, where i,j represents the row & column number of the confidence matrix C, hence obtaining a total score out of 4 for each set of annotated data.

$$C.C.Score(Annotator[i]) = \sum_{j=1}^{5} C(i, j) \quad (1)$$

$$\max(C.C.Score(Annotator[i])) \quad (2)$$

For each facial expression, the annotated dataset corresponding to maximum cumulative confidence score was kept in our final dataset.

### 3.3 Dataset Analysis

While gathering the dataset, the implicit requirement of ethnically diverse India-specific data was the prime focus. Data from both sources, i.e., human subjects and online sources, was gathered with the fore-said requirement as the underlying theme.

**Human Subjects** : Out of 600 human participants 90 were children, 114 belonged to teenagers, 132 were young-adults, 144 belonged to adults, and 120 were from senior-citizens category. The participants were selected from all across the country, with 144 subjects from northern India, 138 from southern India, 114 each from eastern and western zones, and 90 participants belonging to northeastern India. There was about equal participation between males and females with 304 male and 294 female subjects.

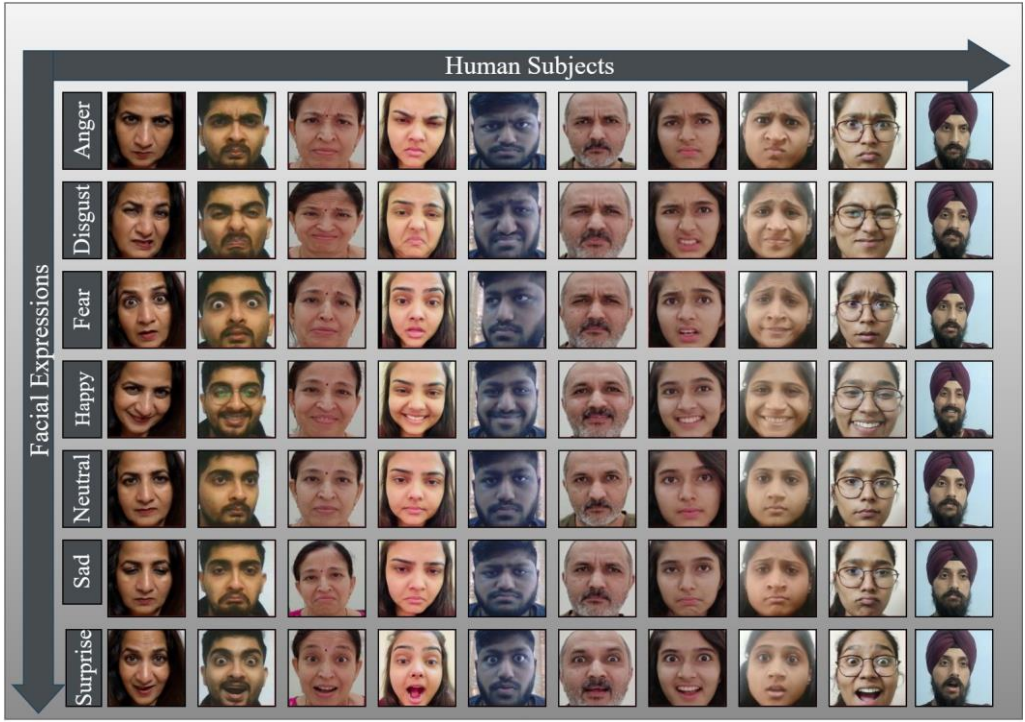

Figure 2: Sample images seven basic expressions captured from human participants.

Fig. 2 shows sample images seven basic expressions captured from human participants. Participants were guaranteed that their identity would be kept confidential and that the captured data would only be used for research purposes. Some of the subjects agreed to use their facial images in research article.

**Online Sources** : Apart from 600 subjects, the proposed dataset also consists of 6000 images crowd-sourced from various online sources using an image search API. Approximately equal representation of both genders was collected, with about 3,120 male (52%) and 2880 female(about 48%) faces. Images from across the five zones were gathered, wherein northern and southern zones had 1500 images (25%) each, eastern and northeastern zones contributed 900 images (15%)each, while the western zone consisted of 1200 images (20%). Fig 3 shows sample images seven basic expressions collected from online sources.

The presence of occlusions disguises some of the crucial FER features, which is an inherent challenge for deploying a FER system for practical applications. Henceforth, some occlusions were included in the dataset, which consisted of spectacles, facial hair, ornaments, turbans, and some unrestricted hand movements while expressing their emotional state

The authors would make the dataset available to those interested upon a reasonable request, they may drop an email to {pratik.narang (or) jagatsesh } @pilani.bits-pilani.ac.in.

## 4 Experimental Results & Analysis

In this section, we discuss the experimentation results on the baseline and state-of-the-art available deep-FER methods on our proposed InFER dataset. The dataset consists of captured human subjects and crowd-sourced online images. Since we already did initial facial detection & extraction for the human subjects, we perform a similar task on images collected from online sources. Haarcascade classifier (Viola and Jones, 2001), was used for detection and extraction of facial images. For a few of the images where Viola & Jones failed, faces were extracted manually. Both the extracted faces (subjects & crowd-sourced ) were combined, and these images were subjected to pose normalization to suppress the effect of in-plane rotations. All the images were resized to a size of 256X256 pixels and converted into grayscale for further processing. 10-fold cross-validation was adopted in our experiment. All the experiments were carried out on an Intel(R) Xeon (R) 2.20 GHz CPU with RAM of 24 GB and a Tesla P100 with 16 GB of graphical memory.

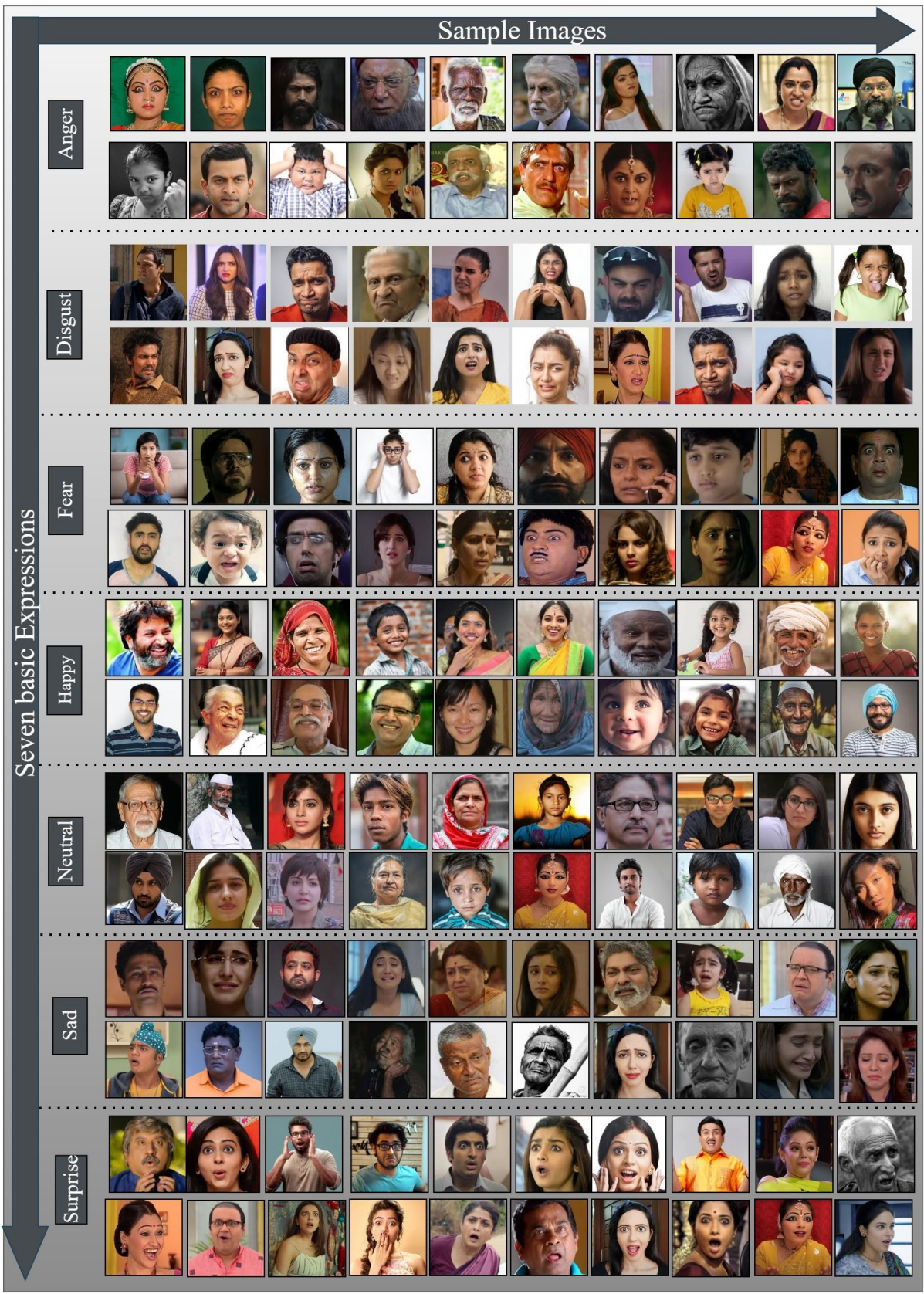

Figure 3: Sample images of seven basic expressions collected from online sources.

Table 3: Traditional & Deep FER accuracies on InFER dataset with 10 fold cross validation. Methods with the highest performance are mentioned in bold, both for Traditional & Deep FER.

| | Method | Classifier | Precision | Recall | F1 Score | Average Accuracy(%) |
|---|---|---|---|---|---|---|
| **Traditional FER** | Grey Intensities(Wang et al., 2010) | PCA + KNN | 0.5520 | 0.5371 | 0.5315 | 54.47 |
| | | **PCA + LDA** | **0.5204** | **0.5181** | **0.5176** | **56.09** |
| | | PCA + Naive-Bayes | 0.4617 | 0.4301 | 0.4139 | 48.78 |
| | LBP Features(Ojala et al., 2002) | $LBP_{8,1}$ from 5x5 region + PCA + LDA | 0.4369 | 0.14907 | 0.4268 | 43.09 |
| | | $LBP_{8,1}$ from 7x6 region + PCA + LDA | 0.5030 | 0.4991 | 0.4974 | 51.22 |
| | | $LBP_{8,1}$ without division + PCA + LDA | 0.5739 | 0.4973 | 0.4926 | 47.15 |
| | | $LBP^{u2}_{8X1}$ from 5x5 region + PCA + LDA | 0.3939 | 0.3945 | 0.3908 | 39.84 |
| | | $LBP^{u2}_{8X1}$ from 7x6 region + PCA + LDA | 0.4060 | 0.4159 | 0.4027 | 41.46 |
| | | $LBP^{u2}_{8X1}$ without division + PCA + LDA | 0.5149 | 0.5116 | 0.5089 | 52.85 |
| | Gabor Filters(Deng et al., 2005) | PCA + LDA | 0.4915 | 0.4701 | 0.4711 | 49.59 |
| | | Adaboost | 0.3511 | 0.3328 | 0.3361 | 39.02 |
| | LGBP(Moore and Bowden, 2011) | $LBP^{u2}_{8X1}$ from 3x3 region + PCA + LDA | 0.6602 | 0.4831 | 0.5219 | 50.40 |
| | | $LBP^{u2}_{8X1}$ from 3x3 region + PCA + LDA | 0.3227 | 0.3287 | 0.3216 | 32.52 |
| | Model | Method | Classifier | Accuracy | | |
| **Deep FER** | **Savchenko et al.** (Savchenko et al., 2022) | **EfficientNET FER** | Softmax / Linear classifier | **78.75** | | |
| | Island Loss(Cai et al., 2018) | IL-CNN | Softmax | 72.8 | | |
| | Center Loss(Cai et al., 2018; Wen et al., 2016) | CL-CNN | Softmax | 71.77 | | |
| | Khorrami et al.(Khorrami et al., 2015) | Zero-Bias CNN | Softmax | 53.97 | | |
| | Pramerdorfer et al.(Pramerdorfer and Kampel, 2016) | Baseline CNN | Softmax | 53.05 | | |

**Traditional FER**: The performance of a FER system immensely depends upon the selection of an appropriate feature extractor. For this purpose, both geometric & appearance-based feature extractors may be used. In our experimentation, have used gray-scale intensities (Wang et al., 2010), local binary patterns (Ojala et al., 2002), gabor wavelets (Deng et al., 2005) and local gabor binary patterns (Senechal et al., 2011) as feature extractors for classifying facial expressions.

**Deep FER** : In recent years, several deep learning-based FER methods have been proposed in the literature. Some state-of-the-art open source methods which were used in our experimentation include EfficientNET FER(Savchenko et al., 2022), Island loss (Cai et al., 2018), and Zero-Bias CNN (Khorrami et al., 2015), and Deep CNN (Pramerdorfer and Kampel, 2016).

Table 3 enlists various traditional & deep FER methods with their respective accuracies on our dataset.

## 4.1 Experimental Results

In the traditional methods, we have first used grayscale intensities of the whole face as the feature extractor. These features were then flattened, and PCA was used to reduce the dimensionality. The number of features was selected in such a way that the original signal could be constructed with an error of less than 5 percent. Subsequently, three classifiers were used; KNN, LDA, and Naive Bayes. For KNN (K-Nearest Neighbors), the number of neighbors was 1. The solver used for LDA (Linear Discriminant Analysis) was SVD (Singular Value Decomposition). The Gaussian variant of the Naive Bayes classifier was used in the experimentation. The PCA + LDA variant gave the best results.

The next feature extractor was LBP. The images were first divided into 5x5 or 3x3 smaller patches. LBP was then applied to extract the features from the patches. The features from patches were then concatenated to form the image again. Experiments were also conducted without dividing the images and directly taking the LBP features. Two variations of LBP were used; default and uniform. Default LBP on the whole face gave the best results.

For extracting features using Gabor filters, the whole image was used. The two classifiers used here were AdaBoost and LDA. PCA was again used for dimensionality reduction before using LDA. The PCA + LDA combination gave the best results.

Finally, we have the LGBP feature extractor, combining the Gabor filter features and the LBP features. We first applied the Gabor filter and then used the LBP method to get the LGBP features. PCA + LDA combination was again used for classification. LGBP features on the 5x5 image patches gave the best results. All the methods were tested using 10-fold cross-validation; detailed results are provided in table 3 (Traditional FER section); precision, recall, F1 scores, and average accuracies are provided for each method used in our experimentation.

In the deep FER methods, five state-of-the-art, open source recent methods were used in our experimentation. Applying deep learning-based techniques clearly showed performance improvement as compared to traditional handcrafted methods. The model by (Savchenko et al., 2022) fetched the best accuracy of 78.75%. All methods used a softmax layer or a linear transformation as the classifier. The models were trained for at least 50 epochs. The respective accuracies are reported in table 3 (Deep FER section) with 10 fold cross validation. These results further substantiate the usability of our dataset in real-world practical applications.

## 5 Conclusion & Future Work

Facial expressions being one of the most prominent ways of conveying one's emotional state and intentions. With the increase in computational capabilities, i.e., Graphical Processing Units (GPUs), more computationally complex deep-learning-based algorithms have been proposed, which have proven to outperform old and existing traditional-handcrafted methods. However, these learning-based approaches substantially depend on the dataset they are trained upon. India being a culturally and ethnically rich country, a home to about 1.4 billion people with various racial identities migrating and settling in the sub-continent. In this context, there existed a need for an India-specific ethnically diverse dataset comprising all seven basic human facial expressions.

The proposed InFER dataset comprises of 10,200 images & 4,200 videos of seven basic facial expressions with their age, gender, and ethnic labels. The subject selection done in this regard corroborated that there should not be any dataset bias with respect to ethnicity, age, class, or gender. Moreover, since posed human expressions lack in realistic data, we adopted a two way collection strategy. Whilst posed expressions from human subjects were captured; on the contrary, we also collected realistic spontaneous/acted expressions collected on a crowd-sourced basis from online sources. As a result, the dataset had a good combination of posed and spontaneous expressions levying the benefits of both lab-based and real-world datasets. We also conducted extensive experimentation on baseline models and available state-of-the-art deep-learning-based models, showing that our propsed dataset can be deployed for real-world practical applications.

The Multi-Ethnic Indian Facial Expression Recognition (InFER) dataset would facilitate researchers to train and validate their algorithms for real-world practical applications.